{

\documentclass[journal]{IEEEtran}

\usepackage{mathtools, nccmath}

\usepackage[dvipsnames]{xcolor}
\usepackage{algorithm}
\usepackage{algpseudocode}
\usepackage{multirow}
\usepackage{amsmath}
\usepackage{mathtools}
\usepackage{graphicx}
\usepackage{xcolor}
\usepackage{enumitem,kantlipsum}

\usepackage[normalem]{ulem}

\usepackage{cancel}
\setlist{nolistsep}

\ifCLASSINFOpdf
\else
\fi
\hyphenation{op-tical net-works semi-conduc-tor}

\newcommand{\FIRSTEA}{\textcolor{black}{EA-p-MCTS}}
\newcommand{\SECONDEA}{\textcolor{black}{EA-MCTS}}
\newcommand{\THIRDEA}{\textcolor{black}{SIEA-MCTS}}

\newtheorem{mydef}{Def.}

 \usepackage{amssymb}
\usepackage{pifont}
\usepackage{tikz}

\usepackage{float,booktabs,array}

\begin{document}
%


\title{Evolving the MCTS Upper Confidence Bounds for Trees Using a Semantic-inspired Evolutionary Algorithm in the Game of Carcassonne}

%
%
%

\author{Edgar Galv\'an$^*${\thanks{$^*$Leading,  corresponding and senior author.}},  Gavin Simpson and Fred Valdez Ameneyro
\thanks{The authors are with the Naturally Inspired Comp. Research Group and with the Dep.
of CS, Maynooth University, Ireland. EG and FVA are also with the Hamilton Institute. edgar.galvan@mu.ie,\{gavin.simpson.2021, fred.valdezameneyro.2019\}@mumail.ie}}

\markboth{}
\markboth{}

%



\maketitle

\begin{abstract}
  Monte Carlo Tree Search (MCTS) is a sampling best-first method to search for optimal decisions. The success of MCTS depends heavily on how the tree is built and the selection process plays a fundamental role in this. One particular selection mechanism that has proved to be reliable is based on the Upper Confidence Bounds for Trees (UCT). The UCT attempts to balance exploration and exploitation by considering the values stored in the statistical tree of the MCTS. However, some tuning of the MCTS UCT is necessary for this to work well. In this work, we use Evolutionary Algorithms (EAs) to evolve mathematical expressions with the goal to substitute the UCT formula and use the evolved expressions in MCTS. More specifically, we evolve expressions by means of our proposed Semantic-inspired Evolutionary Algorithm in MCTS approach (SIEA-MCTS). This is inspired by semantics in Genetic Programming (GP), where the use of fitness cases  is seen as a requirement to be adopted in GP. Fitness cases are normally used to determine the fitness of individuals and can be used to compute the semantic similarity (or dissimilarity) of individuals.  However, fitness cases are not available in MCTS. We extend this notion by using multiple reward values from MCTS that allow us to determine both the fitness of an individual and its semantics. By doing so, we show how \THIRDEA $ $ is able to successfully evolve mathematical expressions that yield better or competitive results compared to UCT without the need of tuning these evolved expressions. We compare the performance of the proposed SIEA-MCTS against MCTS algorithms,  MCTS  Rapid Action Value Estimation algorithms, three variants of the *-minimax family of algorithms, a random controller and two more EA approaches. We consistently show how SIEA-MCTS outperforms most of these intelligent controllers in the challenging game of Carcassonne,  whose state-space complexity is, approx., 10$^{40}$.

\end{abstract}


%
\IEEEpeerreviewmaketitle

\section{Introduction}

\label{sec:introduction}

\IEEEPARstart{M}{onte} Carlo Tree Search (MCTS) is a sampling method for finding \textit{optimal decisions} by performing random samples in the decision space and building a tree according to partial results.  The evaluation function of MCTS relies directly on the outcomes of simulations.  The optimal search tree is guaranteed to be found with infinite memory and computation~\cite{kocsis2006bandit}. However, in more realistic scenarios, MCTS can produce good approximate solutions~\cite{Galvan_EnergyCon_2014,galvan2014heuristic}.


The success or failure of MCTS depends heavily on how the MCTS statistical tree is built. The selection policy, responsible for this, behaves incredibly well when using the Upper Confidence Bounds for Trees (UCT)~\cite{10.1007/11871842_29}. Some conditions are to be met for this to work well, \textcolor{black}{for example, the selection of a child of a given node is based on the exploration/exploitation trade-off. To this end, the UCT expression is normally used, yielding good results. As we shall see in Section~\ref{sec:background}, some of the parameters' values can be changed contributing to a better performing MCTS. However, more sophisticated bounds have been proposed such as the single-player MCTS, adding a third term to the UCT formula and changing the value of a parameter~\cite{DBLP:journals/kbs/SchaddWTU12}. The UCB1-tuned also modifies the UCT expression to reduce the impact of the exploration term~\cite{DBLP:journals/ml/AuerCF02}. Thus, it is evident that while the UCT performs well on a range of problems, its adjustment or modification can have a more positive effect. This motivated us to evolve the selection expression based on the UCT formula.} \textcolor{black}{In this work, we use Evolutionary Algorithms (EAs)~\cite{EibenBook2003} to evolve mathematical expressions that can be used instead of the UCT expression}. \textcolor{black}{ EAs could find better adjustments given its bio-inspired form of working: the lack of both problem-specific preconceptions and biases of the algorithm designer opens up the way to unexpected find near-optimal solutions.}

We propose a Semantic-inspired EA in MCTS (SIEA-MCTS). Our approach is inspired by semantics in Genetic Programming (GP)~\cite{Koza:1992:GPP:138936}, where the use of fitness cases are seen as a prerequisite to promote semantic diversity. The adoption of semantics in GP has increased considerably in the last few years due to multiple scientific studies reporting \textcolor{black}{how it improves GP performance}~\cite{DBLP:conf/gecco/GalvanS19,DBLP:conf/ppsn/LopezMES16,Galvan:ASC:2021,DBLP:journals/gpem/UyHOML11}. The adoption of semantics in MCTS is challenging: MCTS does not use fitness cases. The latter are normally used in GP to determine the fitness values of individuals and to compute the semantic diversity between them. Inspired by this, we extend this notion and use multiple reward values from MCTS simulations that allow us to determine both individuals' fitness values and the semantic diversity between them.

To show how a semantic-based approach makes a positive impact in MCTS, we also adopt two more EAs without semantics. One is partially integrated in MCTS: it uses a single reward value returned by the MCTS roll outs to determine the fitness of individuals. The other EA variant uses these reward values and propagates them in the MCTS statistical tree to determine the fitness value of a candidate solution. We also compare these three EA methods against six variants of the MCTS UCT and six variants of the MCTS Rapid Action Value Estimation (RAVE), three variants of the \mbox{*-minimax} family of algorithms and a random controller. We show how \THIRDEA $ $ outperforms all these controllers in the game of Carcassonne, \textcolor{black}{when the same number of roll out simulations are used for those Monte Carlo-based algorithms. SIEA-MCTS performs similar w.r.t. MCTS when the latter uses seven times more roll out simulations.}

The main contributions of this work are:
\begin{enumerate}
\item \textcolor{black}{We show how it is possible to evolve expressions using a simple EA, \textcolor{black}{($\mu+\lambda$)-ES and GP and a compact population}, where these  are used in lieu of the MCTS UCT formula.}
\item \textcolor{black}{This leads to proposing:  EA partially integrated in MCTS (EA-p-MCTS) and  EA in MCTS (EA-MCTS).}
\item \textcolor{black}{During the evolutionary search, we gather valuable information from the MCTS roll out simulations. From this, we extend the notion of fitness cases-based semantics to reward values-based semantics, by using reward values, a semantic-based method, dubbed Semantic-inspired EA in MCTS (SIEA-MCTS).}
\item \textcolor{black}{Through extensive empirical experiments, we show how SIEA-MCTS outperforms all the methods used in this work or is competitive to fined tuned algorithms.}
\end{enumerate}

\textcolor{black}{As we shall see in the following section, the use of EAs in the evolution of a UCT expression has been done before. In this work, however, we use an EA (($\mu+\lambda$)-ES and GP)  using a dramatically smaller population size compared to other inspiring works, leading to its use in this type of problems/games that require to make a relatively fast decision (1$^{st}$ contribution). We then propose three EA variants to evolve the UCT formula: (a) EA-p-MCTS does not propagate the reward values from a given node to the root of the tree, while (b) EA-MCTS fully incorporates this as normally done in MCTS, (c) SIEA-MCTS uses a semantic-based approach to promote diversity (2$^{nd}$ and 3$^{rd}$ contributions). By taking this step-by-step approach, we show how the evolution of UCT expressions alone is not enough to make a positive impact in MCTS (EA-p-MCTS), but instead this should be done in conjunction with the four steps of MCTS (EA-MCTS). Given the size of the population used in this work, we promote diversity through the use of a semantic-inspired approach (SIEA-MCTS). All these motivations were complemented by carrying out fair comparisons between these EAs approaches and other well-known approaches including MCTS, RAVE and MiniMax (4$^{th}$ contribution). 
}

The rest of this paper is organised as follows. \textcolor{black}{Section~\ref{sec:related} discusses how our work fits within the related work.} Section~\ref{sec:background} provides some background in MCTS, EAs and in the game of Carcassonne. Section~\ref{sec:ai:controllers} discusses in detail the controllers used in this work. Section~\ref{sec:experimental} presents the experimental setup. Section~\ref{sec:results} discusses the results obtained by each of the controllers while  Section~\ref{sec:discussion} discusses the evolved expressions.  \textcolor{black}{Section~\ref{sec:disc:contributions} discusses how the contributions of this work have been achieved}. Section~\ref{sec:conclusions} draws some conclusions and discusses future work.

\section{\textcolor{black}{Related Work}}
\label{sec:related}

\textcolor{black}{There are some interesting works using EAs in MCTS. Cazenave~\cite{Cazenave2007EvolvingMT} used GP to evolve the UCT formula to be used in MCTS. The author used Swiss tournament selection, reproduction and mutation to be applied in the 128 or 256 individuals. Moreover, the author destroyed GP individuals that were not appropriate.  Cazanave used around 13 specialised functions plus basic functions. The author tested his approach using the game of Go. He demonstrated how his GP approach outperformed MCTS and RAVE when a relatively small number of playouts were used, but UCT outperformed his approach when more playouts were allowed.}

\textcolor{black}{Motivated by Cazenave's studies, a decade later, Bravi et al.~\cite{Bravi2017EvolvingGU} evolved the MCTS UCB1 formula tested in the General Video Game AI framework, focusing on five games. The authors used a GP system, using three types of mutations, crossover and elitism. They used a population of 100 GP trees and the initial population started with a seeded UCT individual and 99 random trees. In their studies, they considered three different scenarios: (a) only given access to the same information as UCB1, (b) given access to additional game-independent information, and (c) given access to game-specific information. The authors showed that, in average, Scenario (c) outperformed the rest of the scenarios including the UCB1.}

\textcolor{black}{More recently, Holmg{\aa}rd et al.~\cite{DBLP:journals/tciaig/HolmgardGLT19} used GP to evolve persona-specific evaluation formulae to be used in MCTS instead of the UCB1 in the deterministic game of MiniDungeons 2. The authors used a tree-based GP system using four basic functions as well as constant and variables suitable for the game.  To evolve the 100 GP individuals over 100 generations, the authors used selection, crossover, mutation and elitism, as well as an island model. Holmg{\aa}rd et al. reported that their evolved personas were able to play the game more efficiently compared to the UCB1 agents.}

\textcolor{black}{Other interesting works include those carried out by  Alhejali and Lucas~\cite{6633639}, where they used a  GP system to enhance MCTS during the roll out simulations tested in the game of Ms PacMan. The authors used multiple functions including comparison, conditional, logical operators. They also used different types of terminals including action, local and numerical terminals. They used 100 or 500 individuals evolved, using selection and mutation operators, over 50 or 100 generations, where a single run took around 18 days to finish, for the latter case. Another interesting work is related to handling prohibitive branch factors in MCTS through EAs, as proposed by  Baier and Cowling~\cite{8490403} in the deterministic game of Hero Academy. Lucas et al.~\cite{Lucas2014} used an EA as a source of control parameters to bias the roll outs of MCTS. They showed how their proposed approach significantly outperforms the vanilla MCTS using the Mountain Car benchmark problem and a simplified version of Space Invaders.}

\section{Background}
\label{sec:background}


\subsection{The Mechanics Behind MCTS}

MCTS relies on two elements: (a) that the true value of an action can be approximated using simulations, and (b) that these values can be used to adjust the policy towards a best-first strategy. The algorithm builds a partial tree, guided by the results of previous exploration of that tree. Thus, the algorithm iteratively builds a tree until a condition is reached, then the search is halted and the best performing action is executed.

The most accepted steps involved in MCTS  are: (a) \textit{Selection}: a selection policy is recursively applied to descend through the built tree until an expandable node has been reached (a node is classified as expandable if it represents a non-terminal state and also, if it has unvisited child nodes), (b) \textit{Expansion}: normally one child is added to expand the tree subject to available actions, (c) \textit{Simulation}: from the new added nodes, a simulation is run to get an outcome, and (d) \textit{Back-propagation}: the outcome obtained from the simulation step is back-propagated through the selected nodes.

Simulations in MCTS start from the root state and are divided in two stages: when the state is added in the tree, {a tree policy is used to select the actions}. A default policy is used to roll out simulations to completion, otherwise. 


\subsubsection{Upper Confidence Bounds for Trees}

MCTS works by approximating `real' values of the actions that may be taken from the current state. This is achieved through building a decision tree. The success of MCTS depends heavily on how the tree is built and the selection process plays a fundamental role in this. One particular selection mechanism that has proven to be reliable is the UCB1 tree policy~\cite{10.1007/11871842_29}. Formally, UCB1 is defined as: \useshortskip

\begin{equation}
  UCT = \overline{X}_j + 2K \sqrt{(2 \cdot ln \cdot n)/n_j}
  \label{eq:uct}
  \end{equation}

\noindent where $n$ is the no. of times the parent node has been visited, $n_j$ is the no. of times child $j$ has been visited. $K$ is a constant. 


\subsection{Evolutionary Algorithms}

Evolutionary Algorithms (EAs)~\cite{EibenBook2003} refer to a set of bio-inspired algorithms that use evolutionary principles to build adaptive systems. EAs work with a population of $\mu$-\textit{encoded} (representation of the) potential solutions to a particular problem. Each solution represents a point in the search space, where the optimal solution lies. The population is evolved by means of genetic operators, over a number of generations, to produce better results to the problem.  Each individual is evaluated using a fitness function. The fitness value assigned to each individual in the population probabilistically determines how successful the individual will be at propagating (part of) its code to future generations. 


In this work, we briefly describe the two methods used:

\subsubsection{Evolutionary Algorithm: Genetic Programming (GP)} GP~\cite{Koza:1992:GPP:138936} is a form of automated programming where individuals are randomly created by using  function and terminal sets required to solve a given problem. Multiple types of GP have been proposed in the literature with the typical tree-like structure being the predominant form of GP in EAs.

\subsubsection{Evolutionary Algorithm: Evolution Strategies (ES)} In ES~\cite{Rechenberg10.1007/978-3-642-83814-9_6}, mutation is the main operator whereas crossover is the secondary, optional, operator. Historically, there were two basic forms of ES, known as the ($\mu,\lambda$)-ES and the ($\mu+\lambda$)-ES. $\mu$ refers to the size of the parent population, whereas $\lambda$ refers to the number of offspring that are produced in the following generation before selection is applied. In the former ES, the offspring replace the parents whereas in the latter form of ES, selection is applied to both offspring and parents to form the population in the following generation. 

\subsection{Semantics}



For clarity purposes, we first briefly give some definitions on semantics, based on the first author's work~\cite{DBLP:conf/gecco/GalvanS19}, that will allow us to describe our approach later in Section~\ref{sec:ai:controllers}.

Let $p \in P$ be a program from a language $P$. When $p$ is applied to an input  $in \in I$, $p$ produces an output $p(in)$. 

\begin{mydef}
\indent  \textcolor{black}{The semantic mapping function $s: P \rightarrow S$  maps any program $g$ to its semantics $s(g)$.}
\label{def:general}
\end{mydef}

This means, $\textcolor{black}{s(g_1) = s(g_2) \Longleftrightarrow \forall in \in I : g_1(in) = g_2(in).}$ The semantics specified in Def.~\ref{def:general} has three properties. Firstly, every program has only and only one semantics. Secondly, two or more programs can have the same semantics. Thirdly, programs that produce different outputs have different semantics. Def.~\ref{def:general} is general as it does not specify how semantics is represented. This work is inspired by a popular version of semantics GP where the semantics of a program is defined as the vector of output values computed by this program for an input set (also known as fitness cases). The latter are not available in MCTS. We then extrapolate this idea to the fitness space. Thus, assuming we use a finite set of simulations, as normally adopted in MCTS, we can now define, without losing the generality, the semantics of a program in the simulations.

\begin{mydef}
  The semantics $s(p)$ of a program $p$ is the vector of values from each independent simulation $sim$,
\label{def:semantics}
  \end{mydef}

Thus, we have that the semantics of a program in MCTS is given by $s(p) = [p(sim_1), p(sim_2), \cdots, p(sim_l)]$,  where $l = |I|$ is the number of independent simulations.

Based on Def.~\ref{def:semantics}, we can define the \textit{Sampling Semantics Distance} (SSD) between two programs $(p,q)$. \textcolor{black}{That is, let $P$ = $\{p_1, p_2, ..., p_N \}$ and $Q$ = $\{q_1, q_2, \cdots, q_N \}$ be the sampling semantic of Program 1 ($p_1$) and Program 2 ($p_2$) on the same set of sample points, then the SSD between $p_1$ and $p_2$ is defined as} ${SSD(p,q) = (|p_1-q_1|+|p_2-q_2|+...+|p_N-q_N|)/N}$, where $S_p = \{p_1,...,p_N\}$ and $S_q = \{q_1,...,q_N\}$ are the SS of programs $p$ and $q$ based on simulations.

We are now in position to use the well-known semantic similarity (SSi) proposed by the first author and colleagues~\cite{DBLP:journals/gpem/UyHOML11}. This indicates whether the SSD between two programs lies between a lower bound $\alpha$ and upper bound $\beta$ or not. It determines if two programs are similar without being semantically identical. The SSi of two programs $p$ and $q$ on a domain is formally defined as

\begin{equation}
  \textcolor{black}{  SSi(p,q) = (\alpha < SSD(p,q) < \beta)}
  \label{eq:ssi}
\end{equation}

\noindent where $\alpha$ and $\beta$ are the lower and upper bounds for semantic sensitivity, respectively. In our work, we set these 5 and 10, respectively.

\subsection{The Game of Carcassonne}

\begin{figure}[t] 
    \centering\includegraphics[width=0.5\columnwidth]{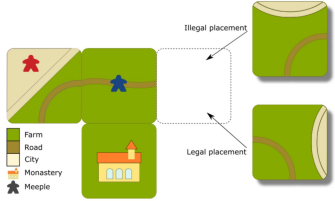}
    \caption{An example of legal and illegal placements of tiles.}
    \label{fig:carcassonne_exmaple}
\end{figure}

The game consists of 72  tiles that are used to build the board. The  tiles contain sections of a landscape (see Fig.~\ref{fig:carcassonne_exmaple}). 

\subsubsection{Placing Tiles}

The game board begins with the same tile being placed face up, and the remaining tiles placed face down in a shuffled deck. In each turn, the player picks up the tile from the deck and must place legally the tile adjacent to any of the previously played tiles on the game board.

\subsubsection{Meeples}

The player can then choose to place one of the seven {meeples} on any of the features in their recently placed tile. The player can only place a {meeple} if they have any left. The player can choose not to place any {meeple} on their turn. The player cannot place a {meeple} on a feature that already contains an opponents' {meeple}. 

\subsubsection{Scoring System}

Scores are added to players' tallies when a feature containing a meeple(s)  is completed. A {city} is finished when the entire section of outer walls are entirely connected. A {city} tile containing a {pennant} symbol is worth double points. To gain scores from a {road}, both ends. {Monasteries} are only scored once they are entirely surrounded by other tiles. If a completed feature contains multiple sets of {meeples}, the points will be awarded to the players with the most {meeples} within the feature. The set of {meeples} are then returned to the players. At the end, {meeples} placed in {farms} gain a score for each completed {city} touching the {farm} it is placed in. Further scores are also added for {meeples} placed in incomplete features. 


\section{AI Controllers}
\label{sec:ai:controllers}

In this work we use seven controllers. A summary of these are: (a) \textbf{MCTS} uses UCT to guide tree search, (b) \textbf{RAVE} memorises, in every node of the tree, the statistics for all possible moves, (c) \textbf{EA-p-MCTS} uses ($\mu$+$\lambda$)-ES and is called during the selection step in MCTS. The fitness of the evolved expression is the score attained in the game of Carcassonne. This value is not backpropagated through the selected nodes, (d) \textbf{EA-MCTS} extends (c) and the fitness of the evolved formula is the result of the average of 30 independent simulations. The value is backpropagated through the selected nodes, (e) \textbf{SIEA-MCTS} extends (d) where selection favours the individual that is semantically different to its parent, (f) \textbf{Minimax} algorithms are expectimax variants that use an alpha-beta pruning techniques, (g) \textbf{Random} actions are executed.

\subsection{Monte Carlo Tree Search}


The core idea of MCTS is based on its four functions: \textit{Selection}, \textit{Expansion}, \textit{Rollout} and \textit{Backpropagation}. The completion of these stages is known as a single simulation. When all simulations conclude, normally the node with the highest action value is chosen, which is the approach taken in this work. The MCTS is explained in detail in Section~\ref{sec:background}.



\subsection{MCTS and Rapid Action Value Estimation}

MCTS-RAVE operates similarly to MCTS. Their main difference is how they treat \textit{action-state} pairs $(s,a)$ in the search tree. With MCTS-RAVE, the value function of a pair $(s,a)$ factors in the average results of simulations when $a$ is selected at any state between the state $s$ and the terminal state. Two new values, $\widetilde{N}(s_t,a_u)$ and $\widetilde{Q}(s_t,a_u)$, are introduced to track the number of times action $a_u$ is selected after state $s_t$ and the average results of these instances, respectively. 

\subsection{Evolutionary Algorithm partially integrated in MCTS}
\label{subsection:EA-p-MCTS}

\begin{algorithm}[tb]
    \small
	\caption{\textcolor{black}{EA-p-MCTS}}
  	\begin{algorithmic}[1]
          \State \noindent \textbf{Input:} Current state \textit{S}, number of generations $G$, lambda $\lambda$\\
        \noindent \textbf{Output:} Evolved tree policy 
	    \Procedure{Evolving\_UCT\_using\_EA-p-MCTS}{$S$, $G$, $\lambda$}
            \State Parent $\gets$ UCT formula
                        \For {$i = 0,  \cdots, G$}
                \State Create $\lambda$ offspring from Parent
                \State Get the fitness of each $\lambda$ offspring using rollouts from $S$
                \State Parent $\gets$ Best performing $\lambda$ offspring
                \EndFor
                \State return Parent
    	\EndProcedure
    	\end{algorithmic}
	\label{alg:mcts_p_es}
\end{algorithm}

We now turn our attention to the proposed AI controller based on EAs to evolve online the \sloppy{mathematical} expression to be used during the selection phase of the MCTS. Alg. 1 shows the steps of our approach. To this end, we use ($\mu$+$\lambda$)-ES (see Section~\ref{sec:background}). We first seed the UCT expression (see Eq.~\ref{eq:uct}) as our initial individual (Line 4). We then proceed to generate the offspring (Line 6). \textcolor{black}{We evolve a candidate solution in every turn that we need to make a decision. At each turn, a new solution is built from scratch. The tree considers and tries to find the best possible opponent move. In the \textit{selection} step of the MCTS algorithm. The reward is multiplied by `-1' when it is an action performed by the opponent. The argmax selection will maximize the opponent’s expected reward.}

Our proposed method aims to evolve  mathematical expressions that can replace UCT  with the goal to get better or competitive results compared to UCT. Thus, ES is called during the selection step in MCTS. Once a  node has been selected by our evolved expression, we proceed to compute the fitness of the evolved expression. We do so by performing roll outs as done in MCTS, but rather than replicating the values from these  until the root node of the MCTS tree and updating the values of the nodes that are in a given path from the tree, we simply keep track of this fitness value to evolve our mathematical expressions. The fitness is the score achieved when selecting a node from the MCTS tree and playing Carcassonne  (Line 7). We dubbed this  method  Evolutionary Algorithm partially integrated in Monte Carlo Tree Search (\FIRSTEA).

\subsection{Evolutionary Algorithm in MCTS}
\label{subsection:EA-MCTS}
           
           
                

\textcolor{black}{We now proceed to describe how the previous  method can be fully integrated in MCTS. We call it Evolutionary Algorithm in Monte Carlo Tree Search (EA-MCTS). The steps of our approach are shown in Alg. 2. We first use the UCT formula as parent (Line 4) to later generate the offspring (Line 8). We assess the fitness of an expression by using the game score coming from the game's state after applying rollouts (Line 11).  The value of these  are used to update a \textit{copy} of the MCTS statistical tree (Line 12), from the selected node to the root including the nodes given in a branch. We perform 30 simulations to compute the fitness of the evolved expression (Line 10). The fitness of our evolved individual is the average of these  30 simulations (Line 15). We pick the best  offspring to act as parent (Line 19).}

\subsection{Semantic-inspired Evolutionary Algorithm in MCTS}
\label{subsection:SIEA-MCTS}

\begin{algorithm}[tb]
    \small
	\caption{\textcolor{black}{EA-SIEA-MCTS}}
  	\begin{algorithmic}[1]
          \State \noindent \textbf{Input:} Number of gen. $G$, lambda $\lambda$, simulations $S$\\
        \noindent \textbf{Output:} Evolved tree policy 
	    \Procedure{Evolving\_UCT\_EA\_SIEA}{}
            \State P $\gets$ UCT formula
           
            \State $T_{copy}$ $\gets$ $T$
           
            \For {$g \gets 0,  \cdots, G$}
                 \For {$i \gets 0, \cdots, \lambda$}
                 \State O$_i$ $\gets$ subtree\_mutation(P)
                 \State a\_fitness $\gets$ 0 
                     \For {$s \gets 0, \cdots, S$}
                        \State tem\_fit(O$_i$) $\gets$ select $T_{copy}(S)$ and rollout(O$_i$) 
                        \State update $T_{copy}$
                        \State a\_fitness(O$_i$) $\gets$ a\_fitness(O$_i$) + temp\_fit(O$_i$)
                     \EndFor
                     \State fitness(O$_i$) $\gets$ a\_fitness(O$_i$) / S
                
                     \EndFor
                     \If {Using EA-MCTS}
                       \State P $\gets$ best offs. 
                     \Else
                        \State  Sem\_Sel(O,P) \textbf{procedure}  
                      \EndIf

             \EndFor
             \State return P
             \EndProcedure

    	\\\hrulefill
\State \noindent \textbf{Input:} Population offspring $O$, Parent $P$\\
        \noindent \textbf{Output:} Best program based on fitness and semantics
             \Procedure{Sem\_Sel}{O,P}
             \State H$_f$ $\gets$ max(fitness($O$))
            \If {More than one offspring from $O$ equals H$_f$}
                    \State SSD $\gets $Sampling\_sem\_dist($O,P$)
                    \State SSi $\gets$ Sel. ind(s) within sem. sim. range ($\alpha$, $\beta$)
                    \If{more than one individual within range}
                        \State New\_P $\gets$ individual closest to lower-bound $\alpha$
                    \Else
                        \State New\_P $\gets$ random($O$)
                    \EndIf
                    \Else
                    \State New\_P $\gets$ random($O$)
            \EndIf
            \State \textbf{return} New\_P
             \EndProcedure
    	\end{algorithmic}
	\label{alg:mcts_siea}
\end{algorithm}

A potential major limitation in using a small population size, as adopted in this work, could be the lack of diversity leading to poor performance. To prevent this, we use semantics as inspiration to promote diversity (see Section~\ref{sec:background}). We dubbed this method Semantic-inspired Evolution Algorithms in Monte Carlo Tree Search (\THIRDEA). We take as basis the EA-MCTS algorithm, shown in Alg. 2 and incorporate semantics into it, shown in the semantic selection procedure of the algorithm, starting in Line 28. 

\textcolor{black}{This controller extends EA-MCTS. The main difference is in the selection of the offspring. We first get the highest fitness, H$_f$, from the offspring (Line 29). If there is more than one offspring with H$_f$ (Line 30), we compute the sampling semantic distance from each offspring with respect to Parent (Line 31). We then proceed to compute the semantic similarity metric using thresholds (Line 32). If there is more than one individual from the offspring population that falls within this threshold, defined by $\alpha$ and $\beta$, then the individual closest to the $\alpha$ value is picked (Line 34). Otherwise, we select an individual randomly from the offspring population (Line 36).}


\subsection{Minimax}

The *-minimax family of algorithms, including  Star1, Star2, and Star2.5, are the most common expectimax variants that use an alpha-beta pruning technique adapted for stochastic trees.  In the Star1 algorithm, the theoretical maximum value $U$ and the theoretical minimum value $L$ are used as the guess for the worst and best scenarios of the chance nodes that have not been evaluated in an attempt to prune the tree if the predicted values fall outside an $\alpha\beta$ window. In the worst-case scenario, no nodes are pruned and the search behaves as  expectimax.

 Star2 is meant for \textit{regular *-minimax games}, in which the actions for each player are influenced by a stochastic event at the beginning of each turn. In Star2, the first node is evaluated and used as the guessed value for the rest of the sister nodes to prune as in Star1, in the {probing phase}. Thus, ordering of the actions is required to get more reliable results and to prune more often. The actions available from each state are ordered as soon as each state is reached for the first time according to how promising they are. If the probing phase fails to achieve a cut-off, the search behaves as the Star1.

 A \textit{probing factor} $f>1$ can be predefined in the Star2.5 algorithm. The probing factor determines the number of nodes to be evaluated during the probing phase. In other words, a $f=\{0,1\}$ refers to  Star1 and Star2, respectively, and $f>1$ refers to Star2.5. We used all these three computationally intensive variants in our studies.

 \subsection{Random}

We also use a controller that chooses moves at random. This controller chooses an action from a set of available possible moves with uniform probability during its turn. This player will predominantly be used as a baseline to demonstrate that the remaining AI controllers are well-informed players.

\section{Experimental Setup}
\label{sec:experimental}

\subsection{Function and Terminal Sets}
The terminal set is defined by  $T =  \{Q(s,a),N(s),N(s,a),K\}$, where $N(s)$ is the number of visits to the node from the MCTS search tree, $N(s,a)$ is the number of visits to a child node, $Q(s,a)$ is the child's node action-value and $K$ is the exploration-exploitation constant. When $K$ is chosen to be mutated, it can take a random value from the following set $r = \{0.25, 0.5, 1, 2, 3, 5, 7, 10\}$. The function set is defined by $F = \{+,-,*,\div,\log,\sqrt { }\}$, where the division operator is protected against division by zero and will return 1 for any divisor less than 0.001. Similarly, the natural log and square root operators are protected by taking the absolute values of input values. The values used for all our controllers are shown in Table~\ref{tab:parameters}.


\begin{table}[tb]
\centering
\caption{Summary of Parameters used in our experiments.}
\resizebox{0.85\columnwidth}{!}{ 
\small\begin{tabular}{|l|r|} \hline 
\emph{Parameter} & \emph{Value} \\ \hline \hline

\multicolumn{2}{|c|}{*-Minimax}\\ \hline
Max Depth & 2 for Star 1, 2 and 2.5 \\ \hline
Lower Bound & $L=-100$, for Star 1, 2 and 2.5 \\ \hline
Upper Bound & $U=100$, for Star 1, 2 and 2.5 \\ \hline
Probing factor & 0, 1, 4, for Star 1, 2 and 2.5, respectively \\ \hline

\multicolumn{2}{|c|}{EA-p-MCTS, ES-MCTS and \THIRDEA}\\ \hline
($\mu$+$\lambda$)-ES &  $\mu=1$, $\lambda=4$  \\ \hline
Generations & 20 \\ \hline
Type of Mutation & \textcolor{black}{Subtree (90\% internal node, 10\% leaf)} \\ \hline
Mutation Rate & One per individual \\ \hline
Initialisation Method & Seeded and mutated  \\ \hline
Maximum depth & 8 \\ \hline
\multicolumn{2}{|c|}{MCTS and MCTS-RAVE}\\ \hline
No. of simulations & 400, 2800 \\ \hline 
K & \textcolor{black}{$\{0.25,0.5,1,2,\sqrt2,3\}$} \\ \hline
$\widetilde{b}$ (RAVE only) &  10\\ \hline


\end{tabular}
}
\label{tab:parameters}
\end{table}

\subsection{League Competition Scoring System}

{Carcassonne} gives a slight advantage to the player that begins the game. Thus, when two controllers face each other, it is important that each gets an equal number of games. A match between two players will consist of a fixed number of games with first and second player not switching between each game. Thus, we will require two matches. This ensures that for a given pair of controllers, each will go as a first player in either Match 1 or Match 2. The result of a match will be decided by the number of wins each player achieves during a match. The two-matches arrangement is analogous to most major professional sport {leagues}, with players/teams playing each other twice in a given season. Inspired by this, we also adopted a score system that allows us to highlight the best controller. Table~\ref{table:points} shows the points given in a game. 

\subsection{Extensive Empirical Experimentation}

To  obtain  meaningful  results,  we  carried  out  an  extensive empirical experimentation: \textcolor{black}{3,990} independent games. This was done as follows. We first group the MCTS controllers to play against each other.  Given that we have \textcolor{black}{six} MCTS variants, \textcolor{black}{we performed 750 independent games (30 matches of 25 games each)}. We carry out the same experiments for the \textcolor{black}{MCTS RAVE (750 independent games)}. We also grouped the three Star algorithms. This led to 150 independent games (six matches of 25 games each).

Finally, we compared our three EA-based MCTS controllers against the best controllers of MCTS, RAVE and Star algorithms. We also considered a random controller and the MCTS and the RAVE controllers with the highest average point difference value. \textcolor{black}{The MCTS and RAVE algorithms each used 400 and 2,800 roll out simulations, independently. The comparison of these controllers only led to 2,340 independent games (156 matches of} 15 games each).



\begin{table}[tb]
  \begin{center}
    \caption{Points awarded to a game of Carcassonne.}
      \resizebox{0.85\columnwidth}{!}{ 

\begin{tabular}{|l|l|r|} \hline
  Acronym     &  Description  & Points\\ \hline \hline

BWP, BLP     & Number of Bonus: Win Points, Loss Points                & 1, 1       \\ \hline
W, L, D       & Number of wins, losses, draws                          & 4, 0, 2      \\   \hline
PD      &  PD=Player Score-Opponent Score &   --     \\  \hline

\end{tabular}
}
\label{table:points}
  \end{center}
  
\end{table}

\section{Discussion of Results}
\label{sec:results}

\begin{table}[tb]
  \begin{center}
    \caption{Number of Wins (W), Losses (L) and Draws (D) by using two different reward systems. Top: R1 using +1, 0 and -1 as reward values for W, D and L, respectively. Bottom: R2 using difference of scores as reward values between the two players. No. of simulations: 400. No. of independent games: 50.}
    \resizebox{0.75\columnwidth}{!}{%
        \begin{tabular}{|l|r|r|r|r|r|r|}
        \hline
         \multirow{2}{*}{Reward system} & \multicolumn{3}{c|}{Player 1}  &   \multicolumn{3}{c|}{Player 2}      \\
            
           & W & L & D & W & L & D  \\    
        \hline \hline
       \textbf{R1.} +1,0,-1   & 2     & 48    & 0     & 4     & 45    & 1                           \\ \hline
       \textbf{R2.}  Diff. of scores       & 45    & 4     & 1     & 48    & 2     & 0                             \\ \hline
        \end{tabular}
        }
    \label{table:mcts:reward}
    \end{center}

\end{table}

\subsection{Reward Systems}

When MCTS is used in a two-player game, the reward value is normally defined as +1, 0 and 1 when the player wins, draws or loses against its opponent, respectively. \textcolor{black}{However, as suggested in~\cite{Fred}, when the difference of scores between the two players is used as reward value is more beneficial.}

\textcolor{black}{Table~\ref{table:mcts:reward} shows the results when using these reward systems: +1, 0, -1 denoted by R1 and  the difference of scores between the two players denoted by R2. We used these controllers to compete against each other using 400 simulations in 50 independent games. When R1 is used as first player, it only wins two games against R2 that is the second player. In the latter case, R2 wins 48 games (see bottom right-hand side of the table). When R2 is used as first player and R1 as second player, we can see that R2 wins 45 games and R1 only wins four games (see top right-hand side of the table). In summary, R2 (difference of scores between the two players to be used as a reward value) yields the best results and is the reward system used in the rest of the experiments.}

\subsection{Performance of MCTS Using Various UCT $K$ Values}

The performance of MCTS is greatly determined by the \textit{exploitation-exploration} constant $K$ in the UCT function (see Eq.~\ref{eq:uct}). Increasing this value will result in less visited nodes or nodes with smaller action-values to be explored more. A lower value for $K$ will closely resemble a greedy algorithm that will exploit the best possible action during each simulation. A round-robin tournament with \textcolor{black}{six} MCTS controllers with different $K$ values (see Table~\ref{tab:parameters} for the values used in this experiment) is adopted in this work to determine what value yields the most competitive MCTS' results.

Table~\ref{table:mcts:matches} shows the results of the matches between pairs of MCTS controllers using different values for $K$ in the UCT. \textcolor{black}{When $K=\sqrt2$ and playing as Player 1 (left-hand side of the table), we can see that this controller wins four matches and loses one: when playing against MCTS $K=1$. When $K=\sqrt2$  plays now as Player 2 (top of the table), it wins three matches, which are against MCTS $K=\{0.25,1,2\}$.} Notice that in some cases the sum of the wins shown in the cells is not equal to 25 games. This denotes a draw.
 
We then use this information to rank each player. This is shown in Table~\ref{table:mcts:results}. Two elements determine this ranking: points attained by a controller (see Table~\ref{table:points}) and in case of a tie, we use  the average point difference between a controller and the rest of the controllers (denoted as PD in Table~\ref{table:mcts:results}).  \textcolor{black}{When MCTS $K=\sqrt2$, this wins seven matches, loses three and there are no draws. This gives 31 points (see second row of Table~\ref{table:mcts:results}). This is computed by considering Table~\ref{table:points}: four points for a win (4 $\times$ 7 wins) and one point for a win/lose with ratio greater than 75:25 (1 $\times$ 3 points).}

\textcolor{black}{We can observe from Tables~\ref{table:mcts:matches} and~\ref{table:mcts:results} that the best performing MCTS controller is when $K=\sqrt2$ is defined in the UCT formula. When $K=\{0.5,2\}$, the controllers show a similar performance with 25 and 22 points, respectively. The last ranked controller ($K=0.25$) achieves 15 points, half of those points obtained by the best performing MCTS controller.}



\begin{table}[tbh!]
  \caption{\textcolor{black}{Number of wins out of 25 games, for each of the 30 pair matches (\textcolor{black}{25$\times$30 = 750 independent games}) when using \textbf{MCTS} and six different $K$ values.}}
    \begin{center}
    \resizebox{0.85\columnwidth}{!}{%
        \begin{tabular}{ll||c|c|c|c|c|c|}
        \cline{3-8}
                                &   & \multicolumn{6}{c|}{{Player 2}}                                       \\ \cline{3-8} 
        \multicolumn{1}{l}{}    &   & $K=0.25$ & $K=0.5$ & $K=1$ & $K=2$ & $K=\sqrt2$ & $K=3$ \\ \hline \hline
        
        \multicolumn{1}{|l|}{\multirow{5}{*}{\rotatebox[origin=c]{90}{{Player 1}}}} 
        
                                        &  $K=0.25$ & -       & 10-14  & 12-12  & 15-10  & 8-17  & 12-13       \\ \cline{2-8} 
        \multicolumn{1}{|l|}{}          &  $K=0.5$  & 11-13   & -      & 15-10  & 15-10  & 13-10 & 15-9       \\ \cline{2-8} 
        \multicolumn{1}{|l|}{}          &  $K=1$    & 9-16    & 11-14  & -      & 15-10  & 10-14 & 17-8        \\ \cline{2-8} 
        \multicolumn{1}{|l|}{}          &  $K=2$    & 16-9    & 19-6   & 13-11  & -      & 11-13 & 16-8        \\ \cline{2-8}
         \multicolumn{1}{|l|}{}          &  $K=\sqrt2$ & 22-3 & 16-9   & 11-13  &   17-7 &  -  & 15-10    \\ \cline{2-8}
        
        \multicolumn{1}{|l|}{}          &  $K=3$    & 17-8    & 14-11  & 10-15  & 12-13  & 13-12 & -           \\ \hline 
        \end{tabular}
    }
    \label{table:mcts:matches}
    \end{center}
    \caption{\textbf{Ranking of the MCTS} controllers based on Points and PD (read text). Table~\ref{table:mcts:matches} shows details of results per match.}

    \begin{center}
    \resizebox{0.85\columnwidth}{!}{%
        \begin{tabular}{|c||l|r|r|r|r|r|r|r|}
          \hline
        {Rank} & {MCTS}  & {Points} & {BWP} & {BLP} & {W} & {L} & {D} & {PD}     \\ \hline \hline
        1 & $K=\sqrt2$ & 31 & 1 & 2 & 7 & 3 & 0 & +40.08 \\ \hline
        2 & $K=0.5$        & 25 & 0 & 1 & 6 & 4 & 0 & +0.44  \\ \hline
        3 & $K=2$          & 22 & 1 & 1 & 5 & 5 & 0 & +7     \\ \hline
        4 & $K=1$          & 19 & 0 & 1 & 4 & 5 & 1 & +22.52 \\ \hline
        5 & $K=3$          & 17 & 0 & 1 & 4 & 6 & 0 & -26.08 \\ \hline
        6 & $K=0.25$       & 15 & 0 & 1 & 3 & 6 & 1 & -43.96 \\ \hline
        \end{tabular}
    }    
    \label{table:mcts:results}
    \end{center}
\end{table}

\subsection{Performance of MCTS-RAVE Using  Various UCT K Values}



\begin{table}[tb]
    \caption{Number of wins out of 25 games, for each of the 30 pair matches (\textcolor{black}{25$\times$30 = 750 independent games}) when using \textbf{MCTS-RAVE} and six different $K$ values.}

    \begin{center}
    \resizebox{0.85\columnwidth}{!}{%
        \begin{tabular}{ll||c|c|c|c|c|c|}
        \cline{3-8}
                                &   & \multicolumn{6}{c|}{{Player 2}}                                       \\ \cline{3-8} 
        \multicolumn{1}{l}{}    &   & $K=0.25$ &  $K=0.5$ &  $K=1$ &  $K=\sqrt2$ & $K=2$ & $K=3$ \\ \hline \hline
        
        \multicolumn{1}{|l|}{\multirow{6}{*}{\rotatebox[origin=c]{90}{{Player 1}}}} 
        
                                        &  $K=0.25$     & -       & 16-9  & 16-9    &  16-9     &23-1   & {25-0}       \\ \cline{2-8} 
        \multicolumn{1}{|l|}{}          &  $K=0.5$      & 14-10   & -     & 19-6    &  20-4     &22-3   & {25-0}       \\ \cline{2-8} 
        \multicolumn{1}{|l|}{}          &  $K=1$        & 8-17    & 6-18  & -       & 17-8      &20-4   & 24-1        \\ \cline{2-8} 
        \multicolumn{1}{|l|}{}          &  $K=\sqrt2$   & 4-21    & 6-18  & 11-13   & -         & 19-6   & 22-3        \\ \cline{2-8} 
        \multicolumn{1}{|l|}{}          &  $K=2$        & {0-25}    & 1-24  & 5-20  & 6-18      &-      & 19-6        \\ \cline{2-8} 
        \multicolumn{1}{|l|}{}          &  $K=3$        & 1-24    & {0-25}  & 2-23  & 4-21      &7-18   & -           \\ \hline 
        \end{tabular}
    }    
    \label{table:mcts:rave:matches}
    \end{center}

        \caption{\textbf{Ranking of the MCTS-RAVE} controllers based on Points and PD (read text). Table~\ref{table:mcts:rave:matches} shows details of results per match.}

    \begin{center}
    \resizebox{0.9\columnwidth}{!}{%
        \begin{tabular}{|c||l|r|r|r|r|r|r|r|}
        \hline
        {Rank} & {MCTS-RAVE}  & {Points} & {BWP} & {BLP} & {W} & {L} & {D} & {PD}     \\ \hline \hline
        1   & $K=0.5$                & 44     & 8   & 0   & 9 & 1 & 0 & +179.48      \\ \hline
        2   & $K=0.25$               & 40     & 4   & 0   & 9 & 1 & 0 & +208.76      \\ \hline
        3   & $K=1$                  & 28     & 4   & 0   & 6 & 4 & 0 & +67.04       \\ \hline
        4   & $K=\sqrt2$             & 21     & 4   & 1   & 4 & 6 & 0 & -18.44       \\ \hline
        5   & $K=2$                  & 9      & 1   & 0   & 2 & 8 & 0 & -154.56       \\ \hline
        6   & $K=3$                  & 0      & 0   & 0   & 0 & 10 & 0 & -282.28      \\ \hline
        \end{tabular}
    }    
    \label{table:mcts:rave:results}
    \end{center}
\end{table}

Let us now turn our attention to the results yield when using MCTS-RAVE. Given that one match between a pair of controllers is composed of 25 games and that each controller acts as first or second player, as well as having  defined \textcolor{black}{six} different $K$ values, we have executed \textcolor{black}{600} independent games (\textcolor{black}{30} matches $\times$ 25 games). The number of wins from this extensive experimentation is shown in Table~\ref{table:mcts:rave:matches}. 

\textcolor{black}{We can see that the best MCTS-RAVE controller is when setting $K=0.5$ (see Table~\ref{table:mcts:rave:matches}). It wins all the matches against all its opponents (five wins) when acting as first player and only loses one match when going as second player when competing against $K=0.25$ (16-9 games). Thus, it wins nine matches (recall one match is equivalent to 25 independent games of Carcassonne). When MCTS-RAVE  $K=0.25$ we can see a similar scenario: it wins all five matches when going as a first player and only loses one match when acting as second player against MCTS-RAVE $K=0.5$. Thus, in total it wins nine out of ten matches. The difference in points between MCTS-RAVE $K=0.5$ and $K=0.25$, ranked first and second, respectively, is due to the fact that the former achieves eight further bonus win points. That is, in eight  matches the ratio of wins is equal or greater to 75:25. Whereas MCTS-RAVE $K=0.25$ achieves only four bonus win points.}

\subsection{Performance of Star1, Star2 and Star2.5}



\begin{table}[tb]
      \caption{Number of wins out of 25 games, for each of the 6 pair matches (25$\times$6 = 150 independent games) when using \textbf{Star1, Star2 and Star2.5}.}

      \begin{center}
        \resizebox{0.5\columnwidth}{!}{%
        \begin{tabular}{ll||c|c|c|}
        \cline{3-5}
                                &   & \multicolumn{3}{c|}{{Player 2}}                                       \\ \cline{3-5} 
        \multicolumn{1}{l}{}    &   & Star1 &  Star2 &  Star2.5  \\ \hline \hline
        
        \multicolumn{1}{|l|}{\multirow{3}{*}{\rotatebox[origin=c]{90}{{\scriptsize{Player 1}}}}} 
        
                                        &  Star1    & -        & 14-11  & 12-12            \\ \cline{2-5} 
        \multicolumn{1}{|l|}{}          &  Star2    & 11-14    & -      & 14-10            \\ \cline{2-5} 
        \multicolumn{1}{|l|}{}          &  Star2.5  & 11-14    & 9-16   & -                \\ \hline
        \end{tabular}
    \label{table:minimax:matches}
}
      \end{center}

    \caption{\textbf{Ranking of the Star} controllers based on Points and PD (read text). Table~\ref{table:minimax:matches} shows details of results per match.}

    \begin{center}
    \resizebox{0.85\columnwidth}{!}{%
        \begin{tabular}{|c||l|r|r|r|r|r|r|r|}
        \hline
        {Rank} & {Player}  & {Points} & {BWP} & {BLP} & {W} & {L} & {D} & {PD}     \\ \hline \hline
        1   & Star1          & 14    & 0   & 0   & 3 & 0 & 1 & +8        \\ \hline
        2   & Star2         & 8     & 0   & 0   & 2 & 2 & 0 & -1.16     \\ \hline
        3   & Star2.5        & 2     & 0   & 0   & 0 & 3 & 1 & -6.84     \\ \hline
        \end{tabular}
    }    
    \label{table:minimax:results}
    \end{center}
\end{table}

Let us now focus our attention on the performance achieved by the Star controllers. Table~\ref{tab:parameters} shows the parameters' values for these controllers. With only three controllers and using a robin-round tournament match, as done with all the experiments reported in this section, we have six matches (one match is composed of 25 games). The results of these matches are shown in Table~\ref{table:minimax:matches}. From this table, we can see that Star1, acting as first player  is able to win a match against Star2 and it attains a draw against Star2.5.  When Star1 is now Player 2,  we can see that this controller wins all the matches (11-14, in both cases, against Star2 and Star2.5). This and the rest of the summary of wins, losses and draws, among other informative values are shown in Table~\ref{table:minimax:results}.  Star1 yields the best results among all the controllers, without losing any matches.

\subsection{Comparison of performance of Evolutionary Algorithms MCTS controllers against the rest of the controllers}

\subsubsection{\textcolor{black}{Difficulties in Fair Performance Comparison}}

\textcolor{black}{The comparison of performance of the algorithms used in this work is problematic: the nature of each of these compared to another one can be radically different.  In this work, we are primarily interested to automatically evolve the UCT formula by means of EAs used in MCTS rather than manually tuning it. Of particular interest is the number of roll out simulations used in our work that allow us to make fair comparisons. To this end, we use 2,800 and 400 roll out simulations for MCTS and MCTS-RAVE. We define  400 roll out  simulations for EA-based MCTS only (EA-p-MCTS, EA-MCTS, SIEA-MCTS). We will discuss in detail the reasons for choosing these number of roll out simulations for these controllers. For completeness, we also report the results attained by Star1 and Random controllers. All these controllers led to 156 matches of 15 games each, leading to 2,340 independent games. The results of this extensive experimentation  are shown in Table~\ref{table:all:matches}.}


  

\begin{table*}[tb]
        \caption{Number of wins out of 15 games, for each of the 156 pair matches (15$\times$156 = 2,340 independent games) when using \textbf{\THIRDEA, \SECONDEA, \FIRSTEA, MCTS $(K=\{0.5,\sqrt2\})$, MCTS-RAVE $(K=\{0.25,0.5\})$, Star1 and Random.}}

  \begin{center}
    \resizebox{0.75\textwidth}{!}{
               \begin{tabular}{llll||c|c|c|c|c|c|c|c|c|c|c|c|c|c|c|c|c|c|c|}
                 \cline{5-17}   &    &&& \multicolumn{13}{c|}{{Player 2}}                              \\
      \cline{5-17}           & & && \multicolumn{3}{c|}{{EA in MCTS}} &  \multicolumn{4}{c|}{{MCTS}} & \multicolumn{4}{c|}{{RAVE}} & \multirow{3}{*}{Star1} & \multirow{3}{*}{Random}  \\
\cline{5-15}  \multicolumn{3}{l}{} & &  \multirow{2}{*}{SIEA} & \multirow{2}{*}{EA} & \multirow{2}{*}{EA-p}  & \multicolumn{2}{c|}{{$K$=0.5}} & \multicolumn{2}{c|}{{$K=\sqrt2$}} & \multicolumn{2}{c|}{{$K$=0.25}} & \multicolumn{2}{c|}{{$K$=0.5}} &     &    \\
\cline{8-15}  \multicolumn{1}{l}{}  & &  &&  &  &  & $2800$ & $400$ & $2800$ & $400$ & $2800$ & $400$ & $2800$ & $400$ &     &     \\\hline \hline
\multicolumn{1}{|l|}{\multirow{13}{*}{\rotatebox[origin=c]{90}{{Player 1}}}} & 
                    \multirow{3}{*}{EA in MCTS} & \multicolumn{2}{|c|}{{SIEA}}  & -      & {10-5}  & 15-0  & 7-8   & 15-0  & 7-8  & 7-6  & 7-7 & 12-3 & 8-7 & 13-2 & 9-6 & 15-0  \\ \cline{3-17} 
\multicolumn{1}{|l|}{}  &  & \multicolumn{2}{|c|}{{EA}}  & {9-6}  & - & 15-0  & 7-8  & 12-3 & 7-8 & 12-3 & 10-4  & 15-0  & 7-8   & 12-3 & 9-6 & 15-0  \\ \cline{3-17} 
\multicolumn{1}{|l|}{}  &  & \multicolumn{2}{|c|}{{EA-p}}   & 2-13   & 2-13  & -     & 0-15  & 3-11  & 0-15 & 0-15  & 3-12  & 3-12 & 1-14 & 3-11 & 1-14 & 15-0  \\ \cline{2-17} 
\multicolumn{1}{|l|}{}  & \multirow{4}{*}{MCTS} & \multicolumn{1}{|c|}{\multirow{2}{*}{$K=0.5$}} & $2800$ & 11-4   & 7-8   & 15-0  & - & 15-0  & 8-6  & 9-6  & 11-4   & 14-1 & 8-7 & 14-0 & 9-5 & 15-0  \\ \cline{4-17} 
\multicolumn{1}{|l|}{}  &  & \multicolumn{1}{|c|}{} & $400$& 5-10   & 8-7  & 15-0  & 4-11   & - & 4-11  & 8-7  & 7-7   & 10-5 & 7-8 & 12-3 & 9-5 & 15-0 \\ \cline{3-17} 
\multicolumn{1}{|l|}{}  &  & \multicolumn{1}{|c|}{\multirow{2}{*}{$K=\sqrt2$}} & $2800$ & 13-2   & 13-2   & 15-0  & 13-2 & 13-2   & -  & 14-1  & 10-5   & 14-1 & 12-3 & 13-2 & 12-3 & 15-0 \\ \cline{4-17} 
\multicolumn{1}{|l|}{}  & & \multicolumn{1}{|c|}{} &  $400$ & 5-9   & 5-10  & 15-0  & 4-11   & 4-11 & 1-13 &- & 5-10  & 8-7   & 4-11 & 12-3 & 7-8 & 15-0  \\ \cline{2-17} 
\multicolumn{1}{|l|}{}  &  \multirow{4}{*}{RAVE} & \multicolumn{1}{|c|}{\multirow{2}{*}{$K=0.25$}} & $2800$ & 7-8   & 9-6  & 14-1  & 4-11   & 10-4   & 5-10 & 9-6 &-  & 14-1   & 8-7 & 11-4 & 10-5 & 15-0  \\ \cline{4-17} 
\multicolumn{1}{|l|}{}  &  & \multicolumn{1}{|c|}{} & $400$  & 2-13   & 2-13  & 13-2  & 2-13  & 4-11  & 1-14   & 8-7 & 5-10 &-    & 3-12  & 7-8 & 8-7 & 15-0  \\ \cline{3-17} 
\multicolumn{1}{|l|}{}  & & \multicolumn{1}{|c|}{\multirow{2}{*}{$K=0.5$}} & $2800$ & 7-8   & 7-8  & 14-1  & 9-5   & 10-5   & 3-12 & 12-3 &10-5 &12-3 &-  & 12-3   & 13-2   & 15-0  \\ \cline{4-17} 
\multicolumn{1}{|l|}{}  &  & \multicolumn{1}{|c|}{} & $400$  & 2-13   & 3-12  & 13-1  & 0-13   & 8-7  & 2-13 & 3-12 & 4-11 & 6-9 & 6-9  & -  & 5-10  & 15-0  \\ \cline{2-17} 
\multicolumn{1}{|l|}{}  &   \multicolumn{3}{|c|}{{Star1}}   & 3-12   & 6-7   & 13-2  & 5-9  & 8-7  & 4-10  & 10-5  & 8-7    & 6-9 & 6-9 & 11-4 &- & 15-0  \\ \cline{2-17} 
\multicolumn{1}{|l|}{}  &   \multicolumn{3}{|c|}{{Random}}  & 0-15   & 0-15  & 0-15  & 0-15  & 0-15  & 0-15  & 0-15  & 0-15  & 0-15  & 0-15  & 0-15  & 0-15  & -     \\ \hline 
\end{tabular}
    }
    \label{table:all:matches}
  \end{center}
    
\end{table*}

\begin{table}[tb]
  \caption{\textbf{Ranking of the \THIRDEA, \SECONDEA, \FIRSTEA, Star1, MCTS ($K=\{1,2\}$), MCTS-RAVE ($K=\{0.25,0.5\}$) and Random} controllers based on Points and PD (read text). Table~\ref{table:all:matches} shows details of results per match.}

    \begin{center}
      \resizebox{\columnwidth}{!}{%

        \begin{tabular}{|c||l|r|r|r|r|r|r|r|}
        \hline
        {Rank} & {Player}  & {Points} & {BWP} & {BLP} & {W} & {L} & {D} & {PD}     \\ \hline \hline
        1   & MCTS ($K=\sqrt2$) (2800)      & 109     & 17   & 0   & 23 & 1  & 0 & +646.6     \\ \hline
        2   & MCTS ($K=0.5$) (2800)         & 94      & 9    & 1   & 21 & 3  & 0 & +420.6     \\ \hline
        3   & \THIRDEA                      & 86      & 10   & 2   & 18 & 5  & 1 & +352.002     \\ \hline
        4   & \SECONDEA                     & 82      & 10   & 4   & 17 & 7  & 0 & +354.267     \\ \hline
        5   & RAVE ($K=0.5$) (2800)         & 82      & 9    & 5   & 17 & 7  & 0 & +352.003     \\ \hline
        6   & RAVE ($K=0.25$) (2800)        & 68      & 5    & 3   & 14 & 8  & 2 & +290.798     \\ \hline
        7   & MCTS ($K=0.5$) (400)               & 54      & 5    & 3   & 11 & 12 & 1 & +131.799     \\ \hline
        8   & Star1                         & 46      & 4    & 2   & 10 & 14 & 0 & +131.198     \\ \hline
        9   & RAVE ($K=0.25$) (400)              & 38      & 4    & 2   & 8  & 16 & 0 & -113.334     \\ \hline
        10  & MCTS ($K=\sqrt2$) (400)            & 38      & 6    & 4   & 7  & 17 & 0 & +121.134     \\ \hline
        11  & RAVE ($K=0.5$) (400)               & 28      & 4    & 0   & 6  & 18 & 0 & -125.535     \\ \hline
        12  & \FIRSTEA                      & 10      & 2    & 0   & 2  & 22 & 0 & -660.333     \\ \hline
        13  & Random                        & 0       & 0    & 0   & 0  & 24 & 0 & -1901.199   \\ \hline
        \end{tabular}
    }    
    \label{table:all:results}
    \end{center}
\end{table}

\subsubsection{\textcolor{black}{Using the same number of roll out simulations for EA-based MCTS, MCTS and MCTS-RAVE}}

\textcolor{black}{Recall that our three EA-based MCTS controllers use 400 roll outs simulations during the MCTS process. We in turn use the same number of simulations for MCTS and MCTS-RAVE. When we focus our attention on Semantic-inspired EA in MCTS (SIEA-MCTS) playing against MCTS ($K=\{0.5,\sqrt2\}$) and MCTS-RAVE ($K=\{0.25,0.5\}$), we can see that SIEA-MCTS wins all the matches (eight wins), regardless of this going as first or second player. The same happens when playing against Star1 and Random controllers (four wins). When SIEA-MCTS plays against EA-MCTS and EA-p-MCTS it wins three matches and loses one against EA-MCTS when the latter plays first. The remaining three wins come when SIEA-MCTS plays against RAVE (2,800 roll out simulations). We discuss the role of this number of simulations later. In total, SIEA-MCTS with 400 simulations, wins 18 matches as summarised in Table~\ref{table:all:results}.}

When we focus our attention on EA-MCTS, we can see that this wins all the matches against MCTS and MCTS-RAVE, \textcolor{black}{ except when EA-MCTS goes second against MCTS $K=0.5$}. Totaling seven wins. EA-MCTS wins three matches against the rest of its EA-variants. It also wins all the matches against Star1 and Random (four wins). It also wins three matches when MCTS and MCTS-RAVE use 2,800 roll out simulations. Specifically, EA-MCTS wins against MCTS-RAVE ($K=0.25$), this as second player, and MCTS and MCTS-RAVE ($K=0.5$, both), these as first players. In total, EA-MCTS wins 17 matches. Table~\ref{table:all:matches} shows the details of these matches. Table~\ref{table:all:results} summarises this.

\textcolor{black}{We can see that SIEA-MCTS and EA-MCTS attain the best results compared to the rest of the controllers, except when MCTS uses seven times the number of simulations that these EA methods use. This is discuss next.}

\subsubsection{\textcolor{black}{Using a higher number of roll out simulations for MCTS and MCTS-RAVE}}

\textcolor{black}{From Table~\ref{tab:parameters}, we can see that we use $\lambda=4$, 20 generations and 30 roll out simulations during the evolutionary process. This gives us 2,400 simulations. Although not all of these simulations are roll out simulations, but a fraction. Every time, the MCTS uses an evolved equation, this uses 400 roll out simulations, giving us a total of 2,800 \textit{evolved and roll out} simulations.}

\textcolor{black}{When MCTS uses 2,800 roll out simulations and EA-based MCTS use 400 roll out simulations (plus 2,400  evolved and roll out simulations), we can see that MCTS wins all four matches  \textit{vs.} SIEA-MCTS, although in two of them by a narrow margin (7-8), when MCTS goes as second player. When MCTS plays against EA-MCTS, the former only wins three matches, from which two are again by a small margin (7-8), and loses one against EA-MCTS (MCTS $K=0.5$ as first player). These marginal wins by manually fined tuned MCTS with 2,800 roll out simulations allow it to be Ranked  1$^{st}$ and 2$^{nd}$ in Table~\ref{table:all:results}.}

\textcolor{black}{When we focus our attention on the performance of EA-based MCTS against MCTS-RAVE (2,800 simulations), we can see that SIEA-MCTS wins three out of four matches, with a draw (MCTS-RAVE $K=$0.25, as second player). Although, it is fair to say that the three wins obtained by SIEA-MCTS are also by a small margin (7-8). When EA-MCTS plays against MCTS-RAVE, the former wins two and loses two matches. Only one of these wins is by a large margin (10-4) and two small margins for a win and a lose.}

\subsubsection{\textcolor{black}{Summary of results}}
  \textcolor{black}{It is difficult to make a fair comparison of our proposed EA-based MCTS \textit{vs} MCTS and MCTS-RAVE given the different nature of these methods. We have, however, made different comparisons to shed some light on the performance of our proposed methods against MCTS and MCTS-RAVE. \textcolor{black}{We have learned that our SIEA-MCTS and EA-MCTS perform much better compared to MCTS and MCTS-RAVE when all these four algorithms use the same number of roll out simulations. They also perform much better compared to Star1, EA-p-MCTS and a Random controller}.}

  \textcolor{black}{This trend continues when MCTS-RAVE uses 2,800 roll out simulations. In this case, SIEA-MCTS and EA-MCTS perform slightly better, even when these two use 400 roll out simulations (plus 2,400 roll out and evolved simulations). This situation, however, changes when we compare the performance of MCTS \textit{vs.} SIEA-MCTS and EA-MCTS. In this case, these two EA-based methods perform slightly worse than MCTS, where the latter marginally wins a few matches. This is amplified in Table~\ref{table:all:results} by considering that each win match is worth four points. However, as we will see in the following section, the results yielded by SIEA-MCTS and MCTS (using 2,800 roll out simulations) are not statistically significant.}

The strength of SIEA-MCTS and EA-MCTS stem from the fact that these algorithms could find better adjustments to the UCT formula  given its bio-inspired form of working: the lack of both problem-specific preconceptions and biases of the algorithm designer opens up the way to unexpected find optimal solutions based on the findings captured during simulations. In stark contrast to this, MCTS and MCTS-RAVE work incredibly well when these are manually well-tuned and enough roll out simulations are used during the sampling process. \textcolor{black}{Moreover, our EA-based MCTS methods, specifically EA-MCTS and SIEA-MCTS, work incredible well for the right problems that require a high number of roll out simulations.}

  \subsection{Statistically Significant Results}

  We used the scores attained by each controller and elaborated a statistical analysis using t-test to determine if a controller is statistically better than another one (the null hypothesis can be rejected at 1\% level). Let us start analysing the results yield by MCTS, shown in Tables~\ref{table:mcts:matches} and~\ref{table:mcts:results}. The best performing controller, $K=\sqrt2$, is statistically significant to $K=0.25$ (ranked last) and not statistically significant to the rest of the MCTS $K$ values.


When we analyse the results of MCTS-RAVE shown in Tables~\ref{table:mcts:rave:matches} and~\ref{table:mcts:rave:results}, we can see that when MCTS-RAVE $K=0.5$ (ranked first), the results obtained by this are statistically significant to RAVE $K=\{1,2,\sqrt2,3\}$ and not statistically significant to $K=0.25$. When we focus our attention on the results obtained by Star algorithms, shown in Tables~\ref{table:minimax:matches} and~\ref{table:minimax:results}, we can see that the best controller, Star1, is not statistically significant to the rest of the Star controllers.

\textcolor{black}{Finally, when we analyse the results of the controllers used in this work (MCTS, MCTS-RAVE each  using independently 2,800 and 400 roll out simulations, referred as S1 and S2, respectively, and each using two different $K$ values as well as Star1 and a Random controller) and our three EA-based MCTS controllers, shown in Tables~\ref{table:all:matches} and~\ref{table:all:results},  we can see that MCTS $K=\sqrt2$ (S1), ranked first, is statistically significant to EA-p-MCTS, EA-MCTS, MCTS $K=\{0.5,\sqrt2\}$ (S2),  MCTS-RAVE $K=\{0.25,0.5\}$ (S2) and MCTS-RAVE $K=0.5$ (S1), Star1 and the Random controller.  MCTS $K=\sqrt2$ (S1) is not statistically significant to SIEA-MCTS, MCTS $K=0.5$ (S1) and MCTS-RAVE $K=0.25$ (S1).}

\begin{figure*}[tbh!]
  \centering
  \begin{tabular}{c}
    \hspace{-0.50cm}  \includegraphics[width=0.87\textwidth]{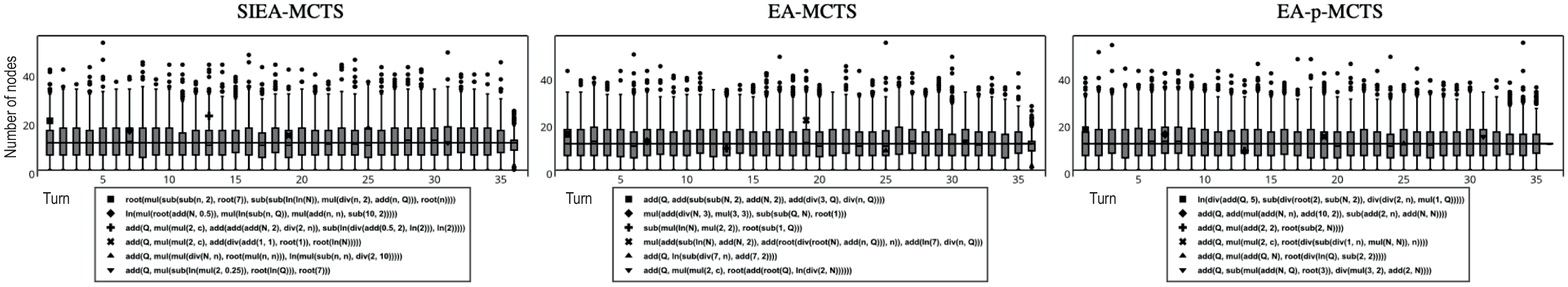}
  \end{tabular}
  \caption{\textcolor{black}{\textit{Top:} Size of evolved expressions measured by number of nodes ($x$-axis) per turn ($y$-axis) for: SIEA-MCTS (left), EA-MCTS (centre) and EA-p-MCTS (right). The horizontal solid line denotes the size of the UCT expression (see Eq.~\ref{eq:uct}). \textit{Bottom:} Examples of evolved expressions taken every `n' turns. Each marker denotes from which turn an expression was taken from during the game (top) and also shows the corresponding expression (bottom).}}

\label{fig:expressions}
\end{figure*}

\section{Evolved  expressions}
\label{sec:evolved}

We keep track of all the evolved expressions generated by our three EA-based proposed approaches when each of these play against MCTS, MCTS-RAVE,  Star1 and Random controllers. The top of Fig.~\ref{fig:expressions} shows the number of nodes ($y$-axis) of these evolved expressions throughout the game of Carcassonne with 36 turns ($x$-axis). The solid horizontal line denotes the size of the UCT expression (see Eq.~\ref{eq:uct}).

\textcolor{black}{Fifty percent of the central data is around the size of the UCT expression (see first and third quartiles of the box plots in Fig.~\ref{fig:expressions}), regardless of the EA method used: SIEA-MCTS (left), EA-MCTS (centre) and EA-p-MCTS (right). This can go as small as e.g., 10 nodes up to +20 nodes. Note how this varies as the game progresses. In particular, it is interesting to observe how at the end of the game (turn 36, right-hand side in each plot of Fig.~\ref{fig:expressions}), SIEA-MCTS and EA-MCTS (the two best performing EA-based MCTS methods) tend to produce expressions of similar length compared to UCT or slightly shorter when the game is about to finish. These fluctuations in the sizes of the evolved UCT formulae, along with the performance discussed in the previous section, indicate how our proposed EA-based MCTS methods uses formulae different to the UCT formula in most of the turns in the game of Carcassonne.}

\textcolor{black}{The bottom of Fig.~\ref{fig:expressions} shows some of the expressions automatically generated by our proposed EA-based MCTS approaches. For illustrative purposes, we show six small expressions picked randomly in Turns 1, 7, 13, 19, 25 and 31, as indicated by six different markers.}

\label{sec:discussion}



\section{Discussion of Contributions}
\label{sec:disc:contributions}

\textcolor{black}{As discussed in Section~\ref{sec:introduction}, we have shown throughout the paper, specifically in Section~\ref{sec:results}, how it is possible to successfully evolve a well-performing selection policy formula in lieu of the well-known UCT using a relatively simple EA using a compact population (1$^{st}$ contribution).}

\textcolor{black}{Furthermore, we have shown in Sections~\ref{sec:background},~\ref{sec:ai:controllers} and ~\ref{sec:results}, step by step what the main elements  are in an EA  that can have a positive effect in the evolution of selection policies, specifically by using EA-p-MCTS and extending this to propose EA-MCTS (2$^{nd}$ contribution). We further extended the latter to propose SIEA-MCTS (3$^{rd}$ contribution). Moreover, ablation studies have been carried out by using this approach. For example, we argue that the use of a semantic-based approach (SIEA-MCTS) can have a more positive effects compared to an EA without promoting a form of semantics (EA-MCTS).}

\textcolor{black}{Finally, we have shown throughout extensive empirical experiments (see, for example, Sections~\ref{sec:results} and~\ref{sec:evolved}), how the proposed SIEA-MCTS is better or competitive to well-known tuned methods such as MCTS UCT $K=\sqrt2$.}




\section{Conclusions and Future Work}
\label{sec:conclusions}

A breakthrough in MCTS was the adoption of the Upper Confidence Bounds for Trees (UCT), which yields extraordinary results provided that this is well calibrated and enough simulations are employed in MCTS. \textcolor{black}{We have shown how when EA-based methods are fully integrated in MCTS, named Semantic-inspired EA in MCTS (SIEA-MCTS) and EA in MCTS, these can have a positive impact in MCTS by using an automatically evolved mathematical expressions to be used in lieu of the well-known fined tuned UCT.}

\textcolor{black}{SIEA-MCTS is particularly interesting since this uses information from roll out MCTS simulations to (i) assess the fitness of evolved individuals and (ii) promote semantic diversity. In the near future, we will carry out research on the optimal values for the lower and upper bounds to be used in the semantic similarity metric. These values determine the success or failure in promoting diversity through semantics in genetic programming. We believe that the same principle carries over in MCTS. Furthermore, in the specialised related literature, most of the studies have focused their attention on the use of a single game. In order to generalise our findings, we will investigate the use of EA-based MCTS in other games }

\section*{Acknowledgments}

This publication has emanated from research conducted with the financial support of Science Foundation Ireland under Grant number 18/CRT/6049. 

\bibliographystyle{abbrv}
\bibliography{eMCTS_arXiv}

\begin{thebibliography}{10}

\bibitem{6633639}
A.~M. Alhejali and S.~M. Lucas.
\newblock Using genetic programming to evolve heuristics for a monte carlo tree
  search ms pac-man agent.
\newblock In {\em 2013 IEEE Conference on Computational Inteligence in Games
  (CIG)}, pages 1--8, 2013.

\bibitem{Fred}
F.~V. Ameneyro, E.~Galv{\'a}n, and {\'A}.~F.~K. Morales.
\newblock Playing carcassonne with monte carlo tree search.
\newblock In {\em 2020 IEEE Symposium Series on Computational Intelligence
  (SSCI)}, pages 2343--2350, 2020.

\bibitem{DBLP:journals/ml/AuerCF02}
P.~Auer, N.~Cesa{-}Bianchi, and P.~Fischer.
\newblock Finite-time analysis of the multiarmed bandit problem.
\newblock {\em Mach. Learn.}, 47(2-3):235--256, 2002.

\bibitem{8490403}
H.~Baier and P.~I. Cowling.
\newblock Evolutionary mcts for multi-action adversarial games.
\newblock In {\em 2018 IEEE Conference on Computational Intelligence and Games
  (CIG)}, pages 1--8, 2018.

\bibitem{Bravi2017EvolvingGU}
I.~Bravi, A.~Khalifa, C.~Holmg{\aa}rd, and J.~Togelius.
\newblock Evolving game-specific ucb alternatives for general video game
  playing.
\newblock In {\em EvoApplications}, 2017.

\bibitem{Cazenave2007EvolvingMT}
T.~Cazenave.
\newblock Evolving monte-carlo tree search algorithms.
\newblock 2007.

\bibitem{EibenBook2003}
A.~E. Eiben and J.~E. Smith.
\newblock {\em Introduction to {E}volutionary {C}omputing}.
\newblock Springer Verlag, 2003.

\bibitem{DBLP:conf/gecco/GalvanS19}
E.~Galv{\'{a}}n and M.~Schoenauer.
\newblock Promoting semantic diversity in multi-objective genetic programming.
\newblock In A.~Auger and T.~St{\"{u}}tzle, editors, {\em Proceedings of the
  Genetic and Evolutionary Computation Conference, {GECCO} 2019, Prague, Czech
  Republic, July 13-17, 2019}, pages 1021--1029. {ACM}, 2019.

\bibitem{Galvan_EnergyCon_2014}
E.~Galv\'an-L\'opez, C.~Harris, L.~Trujillo, K.~R. V\'azquez, S.~Clarke, and
  V.~Cahill.
\newblock Autonomous demand-side management system based on {Monte Carlo} tree
  search.
\newblock In {\em IEEE International Energy Conference (EnergyCon)}, pages 1325
  -- 1332. IEEE Press, 2014.

\bibitem{galvan2014heuristic}
E.~Galv{\'a}n-L{\'o}pez, R.~Li, C.~Patsakis, S.~Clarke, and V.~Cahill.
\newblock Heuristic-based multi-agent monte carlo tree search.
\newblock In {\em IISA 2014, The 5th International Conference on Information,
  Intelligence, Systems and Applications}, pages 177--182. IEEE, 2014.

\bibitem{DBLP:conf/ppsn/LopezMES16}
E.~Galv{\'{a}}n-L{\'{o}}pez, E.~Mezura{-}Montes, O.~A. ElHara, and
  M.~Schoenauer.
\newblock On the use of semantics in multi-objective genetic programming.
\newblock In J.~Handl, E.~Hart, P.~R. Lewis,
  M.~L{\'{o}}pez{-}Ib{\'{a}}{\~{n}}ez, G.~Ochoa, and B.~Paechter, editors, {\em
  Parallel Problem Solving from Nature - {PPSN} {XIV} - 14th International
  Conference, Edinburgh, UK, September 17-21, 2016, Proceedings}, volume 9921
  of {\em Lecture Notes in Computer Science}, pages 353--363. Springer, 2016.

\bibitem{Galvan:ASC:2021}
E.~Galván, L.~Trujillo, and F.~Stapleton.
\newblock Semantics in multi-objective genetic programming.
\newblock {\em Applied Soft Computing}, 115:108143, 2022.

\bibitem{DBLP:journals/tciaig/HolmgardGLT19}
C.~Holmg{\aa}rd, M.~C. Green, A.~Liapis, and J.~Togelius.
\newblock Automated playtesting with procedural personas through {MCTS} with
  evolved heuristics.
\newblock {\em {IEEE} Trans. Games}, 11(4):352--362, 2019.

\bibitem{kocsis2006bandit}
L.~Kocsis and C.~Szepesv{\'a}ri.
\newblock Bandit based monte-carlo planning.
\newblock In {\em European conference on machine learning}, pages 282--293.
  Springer, 2006.

\bibitem{10.1007/11871842_29}
L.~Kocsis and C.~Szepesv{\'a}ri.
\newblock Bandit based monte-carlo planning.
\newblock In J.~F{\"u}rnkranz, T.~Scheffer, and M.~Spiliopoulou, editors, {\em
  Machine Learning: ECML 2006}, pages 282--293, Berlin, Heidelberg, 2006.
  Springer Berlin Heidelberg.

\bibitem{Koza:1992:GPP:138936}
J.~R. Koza.
\newblock {\em Genetic Programming: On the Programming of Computers by Means of
  Natural Selection}.
\newblock MIT Press, Cambridge, MA, USA, 1992.

\bibitem{Lucas2014}
S.~M. Lucas, S.~Samothrakis, and D.~P{\'e}rez.
\newblock Fast evolutionary adaptation for monte carlo tree search.
\newblock In A.~I. Esparcia-Alc{\'a}zar and A.~M. Mora, editors, {\em
  Applications of Evolutionary Computation}, pages 349--360, Berlin,
  Heidelberg, 2014. Springer Berlin Heidelberg.

\bibitem{Rechenberg10.1007/978-3-642-83814-9_6}
I.~Rechenberg.
\newblock Evolution strategy: Nature's way of optimization.
\newblock In H.~W. Bergmann, editor, {\em Optimization: Methods and
  Applications, Possibilities and Limitations}, pages 106--126, Berlin,
  Heidelberg, 1989. Springer Berlin Heidelberg.

\bibitem{DBLP:journals/kbs/SchaddWTU12}
M.~P.~D. Schadd, M.~H.~M. Winands, M.~J.~W. Tak, and J.~W. H.~M. Uiterwijk.
\newblock Single-player monte-carlo tree search for samegame.
\newblock {\em Knowl. Based Syst.}, 34:3--11, 2012.

\bibitem{DBLP:journals/gpem/UyHOML11}
N.~Q. Uy, N.~X. Hoai, M.~O'Neill, R.~I. McKay, and E.~Galv{\'{a}}n-L{\'{o}}pez.
\newblock Semantically-based crossover in genetic programming: application to
  real-valued symbolic regression.
\newblock {\em Genet. Program. Evolvable Mach.}, 12(2):91--119, 2011.

\end{thebibliography}

\end{document}